# Coronavirus (COVID-19) Classification using CT Images by Machine Learning Methods

Mucahid Barstugan[1], Umut Ozkaya[1], Saban Ozturk[2]

Electrical and Electronics Engineering, Konya Technical University[1], Konya, Turkey
Electrical and Electronics Engineering, Amasya University[1], Amasya, Turkey
Corresponding Author: mbarstugan@ktun.edu.tr

**Abstract:** This study presents early phase detection of Coronavirus (COVID-19), which is named by World Health Organization (WHO), by machine learning methods. The detection process was implemented on abdominal Computed Tomography (CT) images. The expert radiologists detected from CT images that COVID-19 shows different behaviours from other viral pneumonia. Therefore, the clinical experts specify that COVİD-19 virus needs to be diagnosed in early phase. For detection of the COVID-19, four different datasets were formed by taking patches sized as 16x16, 32x32, 48x48, 64x64 from 150 CT images. The feature extraction process was applied to patches to increase the classification performance. Grey Level Co-occurrence Matrix (GLCM), Local Directional Pattern (LDP), Grey Level Run Length Matrix (GLRLM), Grey-Level Size Zone Matrix (GLSZM), and Discrete Wavelet Transform (DWT) algorithms were used as feature extraction methods. Support Vector Machines (SVM) classified the extracted features. 2-fold, 5-fold and 10-fold cross-validations were implemented during the classification process. Sensitivity, specificity, accuracy, precision, and F-score metrics were used to evaluate the classification performance. The best classification accuracy was obtained as 99.68% with 10-fold cross-validation and GLSZM feature extraction method.

## 1. INTRODUCTION

COVID-19 disease was occurred in the end of 2019 at Wuhan region of China. COVID-19 disease showed fever, cough, fatigue, and myalgias in human body during early phases (1). The patients had abnormal situations in their CT chest images. The respiratory problems, heart damages, and secondary infection situations were observed as complications of the disease. The findings showed that COVID-19 virus spreads from person to person. The infected person needs to be treated in intensive care unit. The infected people have serious respiratory problems. The CT images of the infected people shows that COVID-19 disease has own characteristics. Therefore, the clinical experts need lung CT images to diagnose the COVID-19 in early phase.

The development of computer vision systems supports the medical applications such as increasing the image quality, organ segmentation, and organ texture classification. The analysis of time series and tumor characteristics (2), the segmentation and detection (3) of tumor modules are some of the machine learning application in biomedical image processing field.

In the literature, there is not a detailed study on coronavirus disease. Xu et al. (4) classified CT images of COVID-19 into three class as COVID-19, Influenza-A viral pneumonia, and healthy cases. They obtained the images from the hospitals in Zhejiang region of China. The dataset consisted of total 618 images, which includes 219 images from 110 patients with COVID-19, 224 images of 224 patients with Influenza-A viral pneumonia, and 175 images of 175 healthy people. They classified the images with 3D-dimensional deep learning model and achieved an 87.6% overall classification accuracy. Shan et al. (5) developed a deep learning based system for segmenting and quantification the infected regions as well as the entire lung on chest CT images. They used 249 COVID-19 patients and 300 new COVID-19 patients for validation in their study. They obtained Dice similarity coefficient as 91.6%. The normal delineation system often takes 1 to 5 hours; however, their proposed system reduced the delineation time to four minutes.

This study used 150 CT images for COVID-19 classification. Before classification process, the four different datasets were created from 150 CT images and the samples of datasets were labelled as coronavirus / non-coronavirus (infected / non-infected). Feature extraction methods and SVM are used during the classification of the coronavirus images. The findings showed that the proposed method could be used to diagnose the COVID-19 disease as an assistant system.

This paper is organized as follows. Section 2 analyses the images statistically and visually. Section 3 briefly explains the feature extraction classification techniques. Section 4 presents the classification results. Section 5 discusses and concludes the results.

## 2. MATERIAL

### 2.1. Statistical Features of Dataset Used

The data consist of 150 CT abdominal images, which belong the 53 infected cases, from the Societa Italiana di Radiologia Medica e Interventistica (6). The patch regions were cropped on 150 CT images. The patches were extracted from the regions selected. Four different patch subsets were created and presented in Table 1.

**Table 1.** Four different subsets created from patch regions

| Subset | Patch Dimension | Number of Non-Coronavirus Patches | Number of Coronavirus Patches |
|---|---|---|---|
| Subset 1 | 16x16 | 5912 | 6940 |
| Subset 2 | 32x32 | 942 | 1122 |
| Subset 3 | 48x48 | 255 | 306 |
| Subset 4 | 64x64 | 76 | 107 |

### 2.2. Visual Features of Dataset

The images in the dataset have acquired from different CT tools. This situation makes the classification process difficult. Because, some grey-levels in one CT image represent the coronavirus infected areas. And the same grey-levels in another CT image represent the non-infected areas. Figure 1 shows the infected areas in images that were acquired from different CT tools.

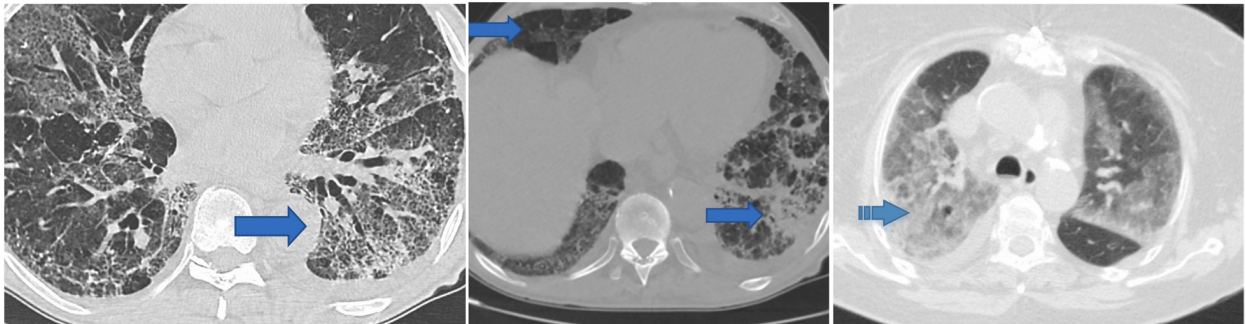

**Figure 1.** The labelled infected areas on CT images

As seen in Figure 1, the grey levels are different in different CT tools. This situation is a disadvantage for classification. Figure 2 shows the patch regions and patch samples from four different subsets.

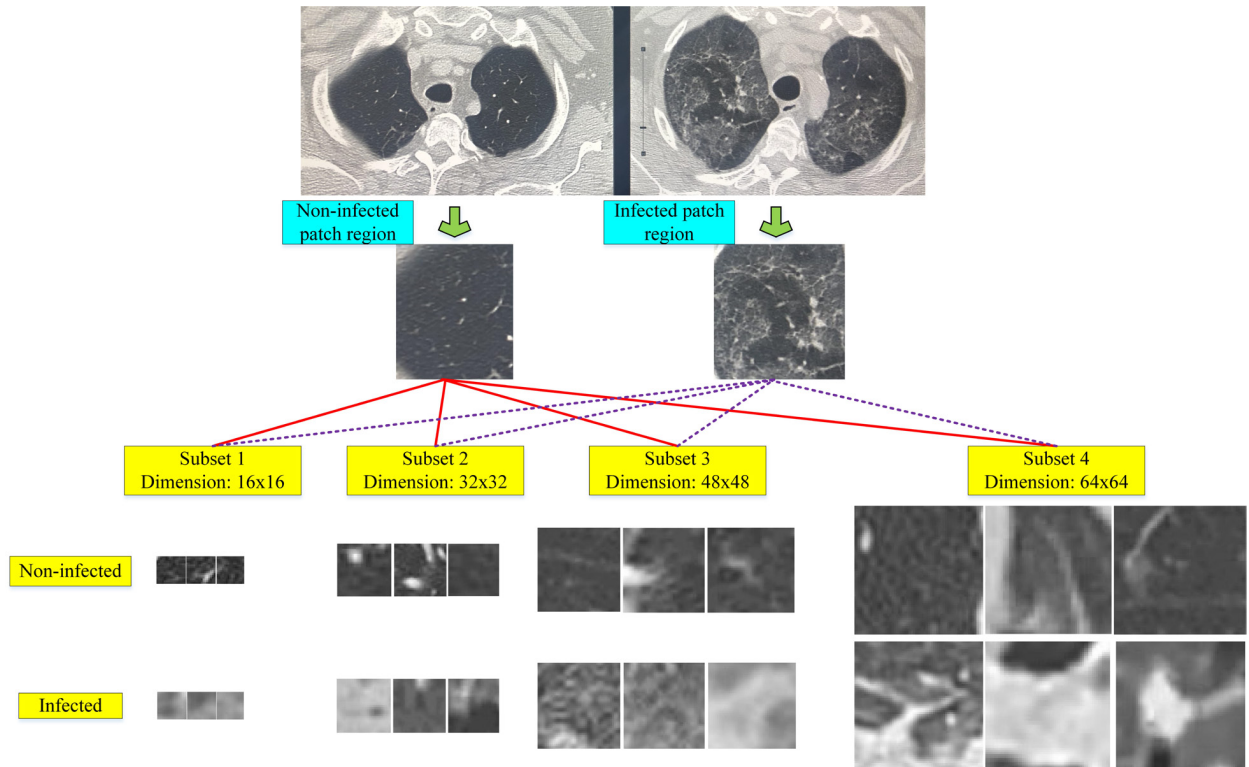

**Figure 2.** Sample images for infected and non-infected situations for all subsets

## 3. METHOD

This study performs a coronavirus classification in two stages. In the first stage, the classification process was implemented on four different subsets without feature extraction process. The subsets were transformed into vector and classified by SVM. In the second stage, five different feature extraction methods such as Grey Level Co-occurrence Matrix (GLCM) (7-9), Local Directional Patterns (LDP) (10), Grey Level Run Length Matrix (GLRLM) (11), Grey Level Size Zone Matrix (GLSZM) (12), and Discrete Wavelet Transform (DWT) (13) extracted the features and the features were classified by SVM (14). During the classification process, 2-fold, 5-fold, and 10-fold cross-validation methods were used. The mean classification results after cross-validations were obtained. Figure 3 shows the two stages of classification process.

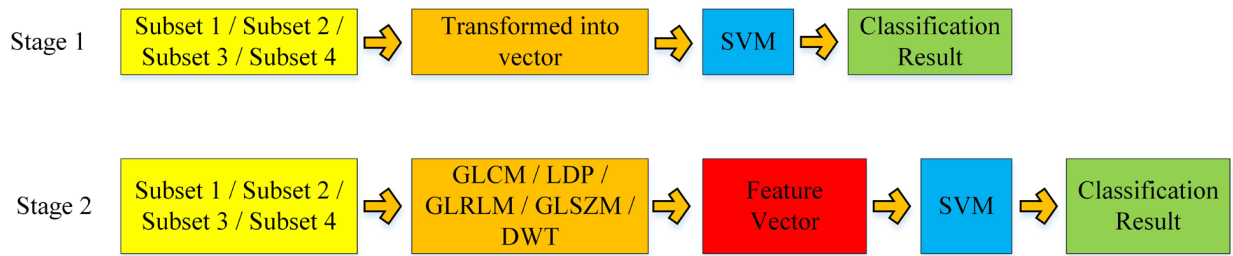

**Figure 3.** The classification process for Stage 1 and Stage 2

## 3.1. The Feature Extraction Techniques

The feature sets formed by using GLCM, LDP, GLRLM, GLSZM and DWT were used for classification of coronavirus. The SVM classifier was used to classify the extracted features, because the SVM is a strong binary classifier. The feature extraction methods used in this study are as follows:

- Grey Level Co-occurrence Matrix
- Local Directional Pattern
- Grey Level Run Length Matrix
- Grey Level Size Zone Matrix
- Discrete Wavelet Transform

### 3.1.1. Grey Level Co-occurrence Matrix

GLCM is used to obtain the second-degree statistical features on the images. GLCM consists of the relationships of different angles between the pixels of an image. Let a co-occurrence matrix that is obtained from an *I* image be represented as *P=[p(i, j | d, Θ)]*. At this point, the co-occurrence matrix is used to evaluate the *i*th pixel frequency features with the *j*th neighbor pixel frequency features by considering the *Θ* direction *and d* length. This study selected *d=1*. And so the *Θ* angle is taken as 0°. GLCM method extracted the *angular secondary moment*, *contrast*, *correlation*, *sum of squares: variance, inverse difference moment, sum average, sum variance, sum entropy, entropy, difference entropy, difference variance, information measures of correlation 1, information measures of correlation 2, autocorrelation, dissimilarity, cluster shade, cluster prominence, maximum probability*, and the *inverse difference* features from all subsets (7-9). GLCM method produces 1x19 feature vector for classifier input.

### 3.1.2. Local Directional Pattern

LDP method uses Kirsch compass kernels to combine the directional elements (30). Let $i_c$ be density of an *I* image on $(x_c, y_c)$. Let $i_n$ be the pixel density when the center pixel $i_c$ is outside of 3x3 neighbourhood of $(x_c, y_c)$. LDP value of $(x_c, y_c)$ is computed as follows (10):

$$YYŞ(x_c, y_c) = \sum_{n=0}^{7} s(i_n - i_c) \cdot 2^n \tag{1}$$

$$s(x) = \begin{cases} 1 & \text{when } x \geq 0 \\ 0 & \text{other} \end{cases} \tag{2}$$

LDP method produces output matrix sized as input image. This matrix is transformed into a vector for classifier input.

### 3.1.3. Grey Level Run Length Matrix

GLRLM extracts texture features on a high level. Let *L* be the number of grey-levels, *R* is the longest run, and *P* is the number of pixels in the image. A GLRLM matrix is *L×R*, and each *p(i,j | θ)* element gives the number of occurrences in the *θ* direction with *i* grey level and *j* run length. GLRLM extracts the *short-run emphasis*, *long-run emphasis*, *grey-level non-uniformity*, *run-length non-uniformity*, *run percentage*, *low grey-level run emphasis*, and *high grey-level run emphasis* features from all subsets (11). GLRLM method produces 1x7 feature vector for classifier input.

### 3.1.4 Grey Level Size Zone Matrix

GLSZM is a feature extraction method, which is developed version of GLRLM algorithm. GLSZM extracts the *small zone emphasis*, *long zone emphasis*, *grey-level non-uniformity*, *size zone non-uniformity*, *zone percentage*, *low grey-level zone emphasis*, *high grey-level zone emphasis*, *small zone low grey-level emphasis*, *small zone high grey-level emphasis*, *large zone low grey-level emphasis*, *large zone high grey-level emphasis*, *grey-level variance*, *and size zone variance* features from all subsets (12). GLSZM method produces 1x13 feature vector for classifier input.

### 3.1.5 Discrete Wavelet Transform

DWT separates the image into frequency sub-bands by using an $h$ low-pass filter and $g$ high-pass filter. Approximation coefficients (LL), horizontal details (LH), vertical details (HL), and diagonal details (HH) represent the lowest frequency, horizontal high frequencies, vertical high frequencies, and high frequencies in both directions, respectively (13). The feature set was created by LL coefficients, which has dimension of the half of input size, after DWT. The LL coefficients were obtained by db1 wavelet, and the coefficient matrix were transformed into a feature vector.

### 3.2. Support Vector Machines (SVMs)

SVM gives high classification accuracy in many applications. An SVM is based on two ideas. The first idea is to map feature vectors to a high dimensional space with a nonlinear method and to use linear classifiers in this new space. The second idea is to separate the data with a high margin hyperplane. This plane is the best plane, which can separate the data as well as possible (14). The cost (C) parameter of SVM algorithm was taken as 1, which is default value of the SVM algorithm for all classification processes.

## 4. EXPERIMENTAL RESULTS

This study presents a coronavirus classification in two stages. Stage 1 classified subsets without feature extraction. Stage 2 implemented feature extraction process on all subsets and classified the features extracted. Five different evaluation metrics (Equations 3-7) were used to assess the proposed method. These metrics are sensitivity (SEN), specificity (SPE), accuracy (ACC), precision (PRE), and F-score.

$$\text{Sensitivity} = TP / (TP+FN) \tag{3}$$
$$\text{Specificity} = TN / (TN+FP) \tag{4}$$
$$\text{Accuracy} = (TP + TN) / (TP + TN + FN+FP) \tag{5}$$
$$\text{Precision} = TP / (TP+FP) \tag{6}$$
$$\text{F-score} = (2*TP)/(2*TP+FP+FN) \tag{7}$$

TP, TN, FP, and FN values are the number of true positives, true negatives, false positives, and false negatives, respectively (15).

*4.1. Classification Results of Subset 1*

Subset 1 has 5912 non-infected and 6940 infected patches. These patches were classified by Stage 1 and Stage 2. Table 2 presents the obtained classification results.

**Table 2.** The classification results for Subset 1

| Stage | Feature Extraction | Number of Features | Cross-validation | Evaluation Metrics (mean (%) ± std) | | | | |
|---|---|---|---|---|---|---|---|---|
| | | | | SEN | SPE | ACC | PRE | F-score |
| Stage 1 | x | 256 | 2-fold | 83.97±2.1 | 76.99±1 | 80.2±0.4 | 75.67±0.4 | 79.6±0.7 |
| | x | 256 | 5-fold | 84.28±1.1 | 77.2±1.6 | 80.5±0.8 | 75.93±1.2 | 79.88±0.7 |
| | x | 256 | 10-fold | 84.23±1.6 | 77.25±1.5 | 80.46±1 | 75.94±1.2 | 79.86±1 |
| Stage 2 | GLCM | 19 | 2-fold | 97.56±0.4 | 98.82±0.3 | 98.19±0.2 | 98.48±0.3 | 98.02±0.2 |
| | GLCM | 19 | 5-fold | 98.41±0.3 | 99.12±0.3 | 98.79±0.3 | 98.96±0.4 | 98.69±0.4 |
| | GLCM | 19 | 10-fold | 98.52±0.3 | 99.23±0.4 | 98.91±0.2 | 99.1±0.4 | 98.81±0.2 |
| | LDP | 256 | 2-fold | 43.72±0.1 | 57.09±0.7 | 50.94±0.3 | 46.47±0.4 | 45.05±0.1 |
| | LDP | 256 | 5-fold | 41.58±2.7 | 58.49±2.3 | 50.71±0.9 | 46.03±1.1 | 43.66±1.8 |
| | LDP | 256 | 10-fold | 42.47±2.5 | 57.71±1.6 | 50.7±1.5 | 46.09±1.8 | 44.19±2.1 |
| | GLRLM | 7 | 2-fold | 98.6±0.1 | 93.67±0.2 | 95.93±0.1 | 92.99±0.2 | 95.71±0.1 |
| | GLRLM | 7 | 5-fold | 98.75±0.5 | 94.29±0.5 | 96.34±0.1 | 93.65±0.5 | 96.1±0.1 |
| | GLRLM | 7 | 10-fold | 98.78±0.5 | 94.38±1 | 96.41±0.6 | 93.75±1.1 | 96.2±0.6 |
| | GLSZM | 13 | 2-fold | 97.34±0.4 | 99.57±0.1 | 98.54±0.2 | 99.48±0.1 | 98.4±0.3 |
| | **GLSZM** | **13** | **5-fold** | **97.56±0.7** | **99.68±0.4** | **98.71±0.3** | **99.62±0.1** | **98.58±0.4** |
| | GLSZM | 13 | 10-fold | 97.72±0.5 | 99.67±0.1 | 98.77±0.2 | 99.6±0.2 | 98.65±0.2 |
| | DWT | 64 | 2-fold | 96.21±0.1 | 98.6±0.3 | 97.5±0.1 | 98.33±0.3 | 97.26±0.1 |
| | DWT | 64 | 5-fold | 96.62±0.7 | 98.62±0.2 | 97.7±0.3 | 98.35±0.2 | 97.47±0.4 |
| | DWT | 64 | 10-fold | 96.8±0.6 | 98.66±0.4 | 97.81±0.3 | 98.4±0.5 | 97.6±0.4 |

As seen in Table 2, the best classification result was obtained as 99.68% in Stage 2 with 10-fold cross-validation and GLSZM feature extraction method.

*4.2. Classification Results of Subset 2*

Subset 2 has 942 non-infected and 1122 infected patches. These patches were classified by Stage 1 and Stage 2. Table 3 presents the obtained classification results.

Table 3. The classification results for Subset 2

| Stage | Feature Extraction | Number of Features | Cross-validation | Evaluation Metrics (mean (%) ± std) | | | | |
|---|---|---|---|---|---|---|---|---|
| | | | | SEN | SPE | ACC | PRE | F-score |
| Stage 1 | x | 1024 | 2-fold | 82.6±0.6 | 77.99±0.6 | 80.09±0.1 | 75.9±0.4 | 79.1±0.1 |
| | x | 1024 | 5-fold | 83.23±2.5 | 80.31±3.1 | 81.64±2.2 | 78.07±2.8 | 80.54±2.3 |
| | x | 1024 | 10-fold | 83.65±2.6 | 79.94±3.2 | 81.64±1.9 | 77.87±2.67 | 80.62±1.9 |
| Stage 2 | GLCM | 19 | 2-fold | 97.55±0.5 | 99.11 | 98.4±0.2 | 98.92 | 98.2±0.2 |
| | GLCM | 19 | 5-fold | 97.77±0.7 | 99.38±0.2 | 98.64±0.4 | 99.24±0.3 | 98.5±0.5 |
| | GLCM | 19 | 10-fold | 97.77±1.6 | 99.47±0.5 | 98.69±0.8 | 99.35±0.6 | 98.55±0.9 |
| | LDP | 1024 | 2-fold | 43.82±1.1 | 54.55±2.3 | 49.66±1.7 | 44.76±1.8 | 44.3±1.4 |
| | LDP | 1024 | 5-fold | 43.1±4.4 | 51.4±2.3 | 47.6±3.8 | 42.7±4.1 | 42.88±4.2 |
| | LDP | 1024 | 10-fold | 42.03±4.9 | 52.49±3.9 | 47.72±1.9 | 42.56±2.4 | 42.22±3.4 |
| | GLRLM | 7 | 2-fold | 54.77±1.2 | 78.7±2.1 | 67.78±0.6 | 68.38±1.7 | 60.81±0.1 |
| | GLRLM | 7 | 5-fold | 58.69±4.9 | 78.25±1.1 | 69.33±2.1 | 69.32±1.8 | 63.5±3.6 |
| | GLRLM | 7 | 10-fold | 60.29±5.4 | 77.99±2.5 | 69.91±2.8 | 69.65±3 | 64.57±4 |
| | GLSZM | 13 | 2-fold | 90.23±0.9 | 94.47±0.3 | 92.54±0.3 | 93.2±0.2 | 91.7±0.4 |
| | GLSZM | 13 | 5-fold | 90.77±2.6 | 94.48±1.8 | 92.78±1.1 | 93.29±1.9 | 91.98±1.3 |
| | GLSZM | 13 | 10-fold | 91.5±1.9 | 95.01±1.5 | 93.41±1.1 | 93.93±1.8 | 92.69±1.2 |
| | DWT | 256 | 2-fold | 98.72±0.3 | 99.55±0.1 | 99.18±0.2 | 99.47±0.2 | 99.09±0.7 |
| | DWT | 256 | 5-fold | 98.94±0.5 | 99.64±0.4 | 99.32±0.3 | 99.57±0.4 | 99.25±0.3 |
| | DWT | 256 | 10-fold | 99.15±1.1 | 99.55±0.8 | 99.37±0.6 | 99.47±0.9 | 99.31±0.7 |

Table 3 shows that the best classification result was obtained as 99.37% in Stage 2 with 10-fold cross-validation and DWT feature extraction method.

### 4.3. Classification Results of Subset 3

Subset 3 has 255 non-infected and 306 infected patches. These patches were classified by Stage 1 and Stage 2. Table 4 presents the obtained classification results.

Table 4. The classification results for Subset 3

| Stage | Feature Extraction | Number of Features | Cross-validation | Evaluation Metrics (mean (%) ± std) | | | | |
|---|---|---|---|---|---|---|---|---|
| | | | | SEN | SPE | ACC | PRE | F-score |
| Stage 1 | x | 2304 | 2-fold | 71.37±0.4 | 70.26±0.5 | 70.77±0.1 | 66.67±0.3 | 68.99 |
| | x | 2304 | 5-fold | 72.94±8.9 | 74.2±4 | 73.62±3.5 | 70.2±3.1 | 71.35±5 |
| | x | 2304 | 10-fold | 74.54±10.5 | 73.55±9.9 | 73.99±7.1 | 70.62±8.9 | 72.19±7.8 |
| Stage 2 | GLCM | 19 | 2-fold | 96.07±2.2 | 99.67±0.5 | 98.04±0.8 | 99.6±0.6 | 97.8±0.9 |
| | GLCM | 19 | 5-fold | 96.47±3.7 | 99.67±0.7 | 98.22±1.7 | 99.61±0.9 | 97.97±1.9 |
| | GLCM | 19 | 10-fold | 96.52±4.9 | 99.67±1.1 | 98.22±2.2 | 99.63±1.2 | 97.98±2.6 |
| | LDP | 2304 | 2-fold | 40.78±0.2 | 55.23±1.4 | 48.66±0.9 | 43.16±0.8 | 41.94±0.5 |
| | LDP | 2304 | 5-fold | 41.57±9.6 | 56.19±5.3 | 49.55±5.4 | 43.86±6.9 | 42.57±8.1 |
| | LDP | 2304 | 10-fold | 39.66±7.6 | 56.6±8.6 | 48.8±4.8 | 43.37±5.9 | 41.14±5.9 |
| | GLRLM | 7 | 2-fold | 53.3±8.6 | 79.74±1.8 | 67.73±2.9 | 68.57±1.6 | 59.85±6.1 |
| | GLRLM | 7 | 5-fold | 56.47±6.9 | 80.08±5.2 | 69.34±3.2 | 70.54±5.6 | 62.48±4.8 |
| | GLRLM | 7 | 10-fold | 57.63±9.6 | 80.12±5.9 | 69.89±5.5 | 70.72±7.6 | 63.22±8 |
| | GLSZM | 13 | 2-fold | 94.5±3.4 | 97.06±2.3 | 95.9±0.3 | 96.49±2.6 | 95.43±0.4 |
| | GLSZM | 13 | 5-fold | 95.29±4.9 | 98.37±1.6 | 96.97±1.5 | 98.08±1.9 | 96.57±1.8 |
| | GLSZM | 13 | 10-fold | 95.26±5.3 | 98.67±2.3 | 97.15±2.3 | 98.52±2.6 | 96.76±2.7 |
| | DWT | 576 | 2-fold | 98.43±1.1 | 99.67±0.5 | 99.11±0.3 | 99.61±0.6 | 99.01±0.3 |
| | DWT | 576 | 5-fold | 98.82±1.1 | 100 | 99.47±0.5 | 100 | 99.41±0.5 |
| | DWT | 576 | 10-fold | 99.23±2.4 | 100 | 99.64±1.1 | 100 | 99.6±1.3 |

Table 4 shows that the best classification result was obtained as 99.64% in Stage 2 with 10-fold cross-validation and DWT feature extraction method.

*4.4. Classification Results of Subset 4*

Subset 3 has 76 non-infected and 107 infected patches. These patches were classified by Stage 1 and Stage 2. Table 5 presents the obtained classification results.

**Table 5.** The classification results for Subset 4

| Stage | Feature Extraction | Number of Features | Cross-validation | Evaluation Metrics (mean (%) ± std) | | | | |
|---|---|---|---|---|---|---|---|---|
| | | | | SEN | SPE | ACC | PRE | F-score |
| Stage 1 | x | 4096 | 2-fold | 86.84±7.4 | 74.74±4.3 | 79.78±0.6 | 71.05±1.5 | 78.04±2.1 |
| | x | 4096 | 5-fold | 78.92±14.5 | 73.94±7.3 | 75.96±3.5 | 68.51±4.1 | 72.74±6.1 |
| | x | 4096 | 10-fold | 72.68±16.8 | 76.6±15.2 | 74.88±9.4 | 70.97±12.7 | 70.51±10.8 |
| Stage 2 | GLCM | 19 | 2-fold | 86.84±3.7 | 97.18±1.4 | 92.9±0.7 | 95.71±1.8 | 91.02±1.2 |
| | GLCM | 19 | 5-fold | 86.83±14.1 | 98.18±2.5 | 93.47±6.2 | 96.93±4.3 | 91.18±8.9 |
| | GLCM | 19 | 10-fold | 86.79±8.9 | 98.1±4 | 93.45±3.5 | 97.64±4.9 | 91.53±4.8 |
| | LDP | 4096 | 2-fold | 31.58±11.2 | 68.23±2.2 | 52.99±3.5 | 40.7±6.7 | 35.41±9.7 |
| | LDP | 4096 | 5-fold | 38.08±10.5 | 66.28±10.3 | 54.67±6 | 45±6.7 | 40.74±9.7 |
| | LDP | 4096 | 10-fold | 33.04±19.6 | 64.6±11.6 | 51.32±12.5 | 38.85±19 | 35.09±18.9 |
| | GLRLM | 7 | 2-fold | 50±3.7 | 80.33±6.9 | 67.75±2.6 | 64.92±6.1 | 56.29±0.1 |
| | GLRLM | 7 | 5-fold | 48.58±11.4 | 80.52±7.1 | 67.28±8.5 | 63.64±14.6 | 55.08±12.7 |
| | GLRLM | 7 | 10-fold | 47.86±18.6 | 79.36±8.6 | 66.17±11.1 | 61.75±15.6 | 53.25±16.4 |
| | GLSZM | 13 | 2-fold | 90.79±1.9 | 84.05±9.5 | 86.87±4.7 | 80.82±8.9 | 85.33±4.2 |
| | GLSZM | 13 | 5-fold | 89.58±7.2 | 92.47±7.9 | 91.29±5.8 | 90.44±9.9 | 89.71±6.2 |
| | GLSZM | 13 | 10-fold | 87.14±14.8 | 94.45±8.8 | 91.22±10.1 | 91.9±12.4 | 89.05±12.6 |
| | DWT | 1024 | 2-fold | 93.42±1.9 | 94.41±2.6 | 93.99±3.5 | 92.24±3.5 | 92.82±2.7 |
| | DWT | 1024 | 5-fold | 93.42±4.7 | 98.18±2.5 | 96.17±1.6 | 97.41±3.5 | 95.26±1.9 |
| | **DWT** | **1024** | **10-fold** | **93.39±7** | **100** | **97.28±2.9** | **100** | **96.46±3.7** |

Table 5 shows that the best classification result was obtained as 97.28% in Stage 2 with 10-fold cross-validation and DWT feature extraction method.

Table 2, Table 3, Table 4 and Table 5 show that the best performance was obtained by extracting features on patches. GLCM, GLSZM and DWT methods always had classification accuracy over 90% during 10-fold cross validation. The best classification performance was achieved by using GLSZM method with 5-fold cross-validation. The scheme of the best method is presented in Figure 4.

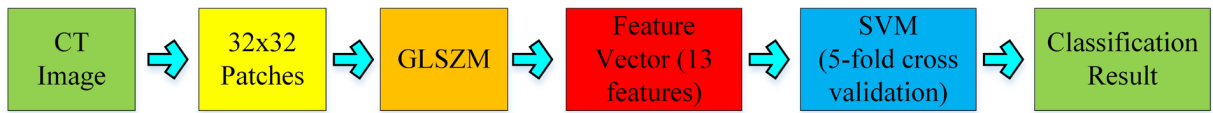

**Figure 4.** The optimum classifier structure for detection of the infected patches

As seen in Figure 4, the CT image was divided into 32x32 sized patches. GLSZM method extracts the features of the patches and form feature vector. The vector is classified by five different SVM structures, which were obtained during training phase. The mean classification performance is obtained by SVM classification.

## 5. DISCUSSION and CONCLUSION

COVID-19 was firstly encountered at Wuhan region in China and have been threatening the public health, trade and world economy. The virus shows the partially similar behaviours with other viral pneumonia. Therefore, the spreading rate of the virus made the situation difficult to be under control. CT imaging results of COVID-19 show that different findings according to other clinical studies. Some situations such as the bronchiectasis, lesion swelling symptoms, and different shadowiness in CT images provide to diagnose COVID-19, easily.

In this study, the coronavirus image set has different type of images, which were acquired with different CT tools. Therefore, five feature extraction methods were utilized to find the feature set that separates the infected patches with a high accuracy. The dataset in this study was formed manually and achieved 99.68% classification accuracy. The proposed method should be tested on another coronavirus CT image dataset.

The literature studies are mostly medical studies. The classification, segmentation studies may increase on COVID-19 in the literature. This study examined COVID-19 images in the classification field. There should be done more classification and segmentation studies on COVID-19. For this aim, the dataset diversion needs to be increased. The machine learning methods should be implemented more on CT abdominal images, X-ray chest images, blood test results when these data were shared to literature.